\documentclass[11pt]{article}

\usepackage[preprint]{acl}

\usepackage{times}
\usepackage{latexsym}
\usepackage{enumitem}

\usepackage[T1]{fontenc}

\usepackage[utf8]{inputenc}

\usepackage{microtype}

\usepackage{inconsolata}

\usepackage{graphicx}

\usepackage{pgfplots}
\usepackage{tikz}

\usepackage{amsmath,amssymb}
\usepackage{algorithm}
\usepackage{algorithmic}
\usepackage{graphicx}
\usepackage{multirow}
\usepackage[table,xcdraw]{xcolor}
\usepackage{subcaption}
\usepackage[normalem]{ulem}
\useunder{\uline}{\ul}{}

\usepackage{pgfplots}
\usepackage{subcaption}
\pgfplotsset{compat=1.17}
\definecolor{datasetA}{RGB}{31,119,180}
\definecolor{datasetB}{RGB}{255,127,14}
\definecolor{datasetC}{RGB}{44,160,44}

\newcommand{\stitle}[1]{\vspace{1.5ex}\noindent{\bf #1}}

\title{SentGraph: Hierarchical Sentence Graph for Multi-hop Retrieval-Augmented Question Answering}

\author{
 \textbf{Junli Liang\textsuperscript{1}},
 \textbf{Pengfei Zhou\textsuperscript{1}},
 \textbf{Wangqiu Zhou\textsuperscript{2}},
 \textbf{Wenjie Qing\textsuperscript{1}},
\\
 \textbf{Qi Zhao\textsuperscript{1}},
 \textbf{Ziwen Wang\textsuperscript{1}},
 \textbf{Qi Song\textsuperscript{1}}\thanks{Corresponding authors},
 \textbf{Xiangyang Li\textsuperscript{1}},
\\
 \textsuperscript{1}University of Science and Technology of China,
 \\
 \textsuperscript{2}Hefei University of Technology
\\
\small{
\texttt{\{jlliang,pengfeizhou,qingwenjie,wangziwen\}@mail.ustc.edu.cn; }} 
\\
\small{
\texttt{rafazwq@hfut.edu.cn; zq2020email@163.com;}}
\small{
\texttt{\{qisong09,xiangyangli\}@ustc.edu.cn}}
}

\begin{document}
\maketitle
\begin{abstract}

Traditional Retrieval-Augmented Generation (RAG) effectively supports single-hop question answering with large language models but faces significant limitations in multi-hop question answering tasks, which require combining evidence from multiple documents. 
Existing chunk-based retrieval often provides irrelevant and logically incoherent context, leading to incomplete evidence chains and incorrect reasoning during answer generation. 
To address these challenges, we propose SentGraph, a sentence-level graph-based RAG framework that explicitly models fine-grained logical relationships between sentences for multi-hop question answering. 
Specifically, we construct a hierarchical sentence graph offline by first adapting Rhetorical Structure Theory to distinguish nucleus and satellite sentences, and then organizing them into topic-level subgraphs with cross-document entity bridges. 
During online retrieval, SentGraph performs graph-guided evidence selection and path expansion to retrieve fine-grained sentence-level evidence. 
Extensive experiments on four multi-hop question answering benchmarks demonstrate the effectiveness of SentGraph, validating the importance of explicitly modeling sentence-level logical dependencies for multi-hop reasoning.

\end{abstract}

\section{Introduction}\label{sec1}

Large Language Models (LLMs) have demonstrated strong capabilities in semantic understanding and text generation, showing broad potential in document reading comprehension tasks~\citep{ application1, llms_development, application2}. 
However, LLMs remain constrained by their internal knowledge boundaries and are prone to hallucination~\citep{Hallucination2, Hallucination1}, particularly in scenarios requiring strict factual accuracy.

To address these issues, Retrieval-Augmented Generation (RAG) incorporates external knowledge to support LLM generation~\citep{rag0, rag1}. 
Traditional RAG methods typically adopt a ``chunk-index-retrieval'' paradigm, retrieving fixed-length text chunks based on semantic similarity~\citep{rag00, rag2}. 
While effective for single-hop question answering (QA), such methods struggle with multi-hop question answering, which requires aggregating evidence across multiple documents, often failing to capture complete evidence chains and leading to incorrect answers.~\citep{HotpotQA, 2wiki, rag_single}.

\begin{figure}[t]
\centering
\includegraphics[width=0.99\columnwidth]{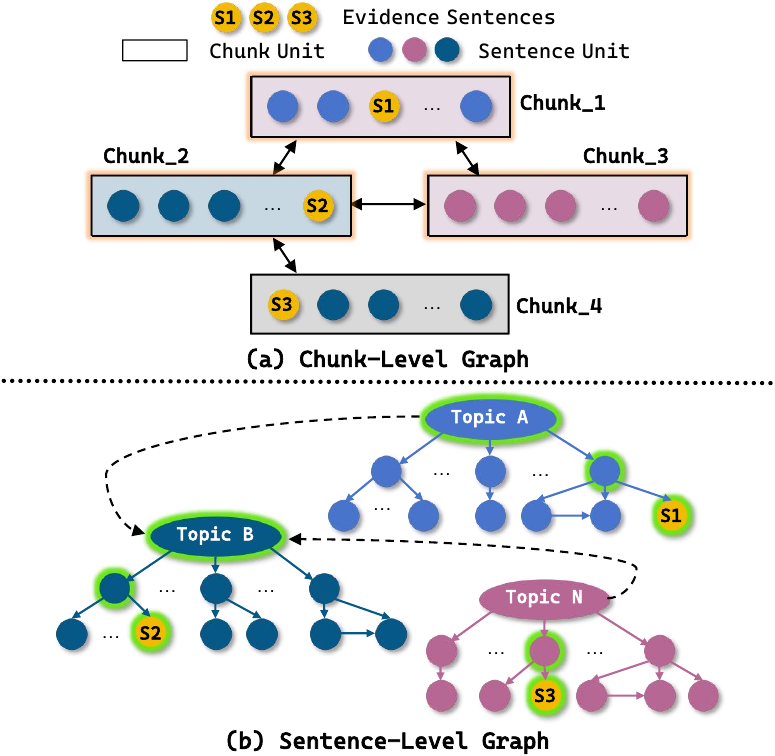}
\caption{Comparison of the traditional chunk-level and our adopted sentence-level graph construction methods.}
\label{fig1}
\end{figure}

To address the limitations of single-step retrieval, several studies explore iterative retrieval strategies to construct evidence chains through multiple retrieval steps~\citep{ircot, RAPTOR}. While these methods can expand evidence coverage, repeated retrieval and reasoning introduce significant latency and computational overhead~\citep{KiRAG}, limiting their use in latency-sensitive applications. Under this constraint, researchers have explored methods to improve the quality of evidence within a single retrieval step.

Post-retrieval optimization methods aim to refine retrieved results to improve their quality~\citep{refiner, RECOMP}. However, they heavily depend on the accuracy of the initial retrieval and often ignore relationships between evidence across different documents, making it difficult to recover missing key evidence~\citep{Shifting}.

To address this limitation, some recent approaches model cross-document relationships by constructing graph-based knowledge structures offline~\citep{KGP, graphRAG, lightrag}. By organizing textual information into structured graphs, these methods facilitate multi-hop evidence capture during retrieval. 
Although such preprocessing introduces additional offline computational overhead, it is performed only once during the knowledge organization stage before deployment. As a result, investing computational resources in offline structured indexing is often a practical design strategy in many retrieval systems.

Nevertheless, existing graph-based retrieval methods still face several limitations. As shown in Figure~\ref{fig1}(a), many approaches primarily rely on similarity-based connections at the chunk level, which makes it difficult to capture fine-grained semantic and logical relationships between key sentences. During retrieval, returned chunks often contain many sentences irrelevant to the query. This redundant information consumes valuable context space and may interfere with downstream reasoning, increasing the risk of hallucination~\citep{irrelevant2}. In addition, weakly relevant but crucial evidence sentences may be overlooked because they are not directly similar to the query, which can lead to incorrect answers.

Inspired by these observations, we propose a sentence graph-based retrieval framework for RAG. Specifically, we reduce the retrieval granularity from chunks to individual sentences and explicitly model the semantic and logical relationships between sentences using a graph structure. However, constructing sentence-level graphs faces several technical challenges:
\emph{\textbf{(1) Context loss}}: Sentences containing anaphoric expressions such as pronouns and deictic terms lose their specific referents when isolated from chunk context, which can lead to ambiguity in understanding the entities or concepts they refer to.
\emph{\textbf{(2) Relationship modeling complexity}}: Unlike chunk-level connections that often rely on surface-level similarity, sentence-level relationships are far more diverse and complex, encompassing various logical types such as causality, condition, and contrast. This diversity makes it challenging to determine which relationship types should be modeled and how to establish these relationships between sentences.
\emph{\textbf{(3) High computational overhead}}: The number of sentences in documents is significantly larger than the number of text chunks. Directly modeling relationships between all sentence pairs would lead to substantial computational cost, making such naive approaches impractical for large-scale corpora.

To overcome these challenges, we propose a hierarchical sentence graph construction framework, as shown in Figure~\ref{fig1}(b). 
To address context loss and relationship modeling complexity, we employ a refined set of relations based on Rhetorical Structure Theory (RST)~\citep{rst}. 
Specifically, we adapt RST by consolidating frequently occurring relations and removing rare ones, resulting in a practical relationship taxonomy that naturally distinguishes between nucleus sentences and satellite sentences. 
To tackle high computational overhead, we design a three-layer graph structure with topic, core sentence, and supplementary sentence layers. 
Concretely, we first construct topic-level subgraphs within documents, then establish cross-document connections through entity-concept bridges at the topic layer.  
In the online retrieval stage, we introduce a sentence graph-based RAG strategy that enables fine-grained evidence selection, thereby reducing irrelevant context and token consumption during LLM generation.

We summarize our contributions as follows:

\begin{itemize}
\item We are the first to apply Rhetorical Structure Theory to sentence-level graph construction for retrieval-augmented generation, providing a principled approach to model fine-grained logical relationships between sentences.
\item We propose an offline hierarchical sentence graph construction method with dual logical relationship modeling, along with an online sentence-level retrieval strategy that leverages the graph structure to retrieve key evidence sentences with their logical context for multi-hop question answering tasks.
\item Extensive experiments on multi-hop question answering benchmarks demonstrate the superior performance of our approach and validate the effectiveness of our framework.
\end{itemize}
\section{Related Work}

\subsection{Retrieval-Augmented Generation}

Retrieval-Augmented Generation has shown strong performance in question answering tasks~\citep{rag0, rag00}.
However, although standard RAG models perform well on single-hop questions, they face clear challenges on multi-hop QA tasks~\citep{HotpotQA}.

To improve retrieval quality, IRCoT\citep{ircot} alternates between chain-of-thought reasoning and knowledge retrieval. The reasoning process guides retrieval, while retrieved evidence is used to refine reasoning. 
KiRAG~\citep{KiRAG} further introduces iterative retrieval based 
on knowledge triples. It explicitly integrates reasoning into the retrieval process to adapt to changing information needs.
While these approaches improve evidence coverage, multi-round retrieval introduces unavoidable computational overhead, limiting their use in latency-sensitive applications.

Under this constraint, post-retrieval optimization methods aim to refine 
retrieved results within a single retrieval step.
Refiner~\citep{refiner} extracts query-relevant content and reorganizes it in a structured form, which helps LLMs better align with the original context.
\citep{Shifting} shifts retrieval from ranking individual passages to 
optimizing the overall quality of a passage set.
However, the performance of these methods still depends heavily on the quality of the initial retrieval results.

\subsection{Graph-Based RAG Methods}

Graph-based methods model cross-document relationships by constructing knowledge structures offline, facilitating multi-hop evidence capture within a single retrieval step.
KGP \citep{KGP} constructs a knowledge graph over multiple documents and designs an LLM-based graph traversal agent.
This agent supports cross-document retrieval and question answering.
GraphRAG \citep{graphRAG} introduces a graph-based approach that leverages LLMs to extract entities and relationships from documents, constructing a knowledge graph that captures semantic structure. Through community detection, it generates hierarchical summaries at multiple levels, enabling both local and global reasoning for complex queries. 
LightRAG \citep{lightrag} proposes a graph-enhanced retrieval framework. Instead of relying on community detection and hierarchical summarization, it adopts a dual-level retrieval strategy that enables both low-level and high-level knowledge discovery.

However, most existing graph-based methods operate at the chunk level.
They treat multi-sentence text chunks as graph nodes, which limits their ability to capture fine-grained semantic relations.
In addition, retrieved chunks often contain irrelevant sentences, introducing noise and increasing the risk of missing weakly related but critical evidence due to coarse chunk-level similarity.

\section{The Proposed SentGraph Method}

\begin{figure*}[ht]
\centering
\includegraphics[width=0.95\textwidth]{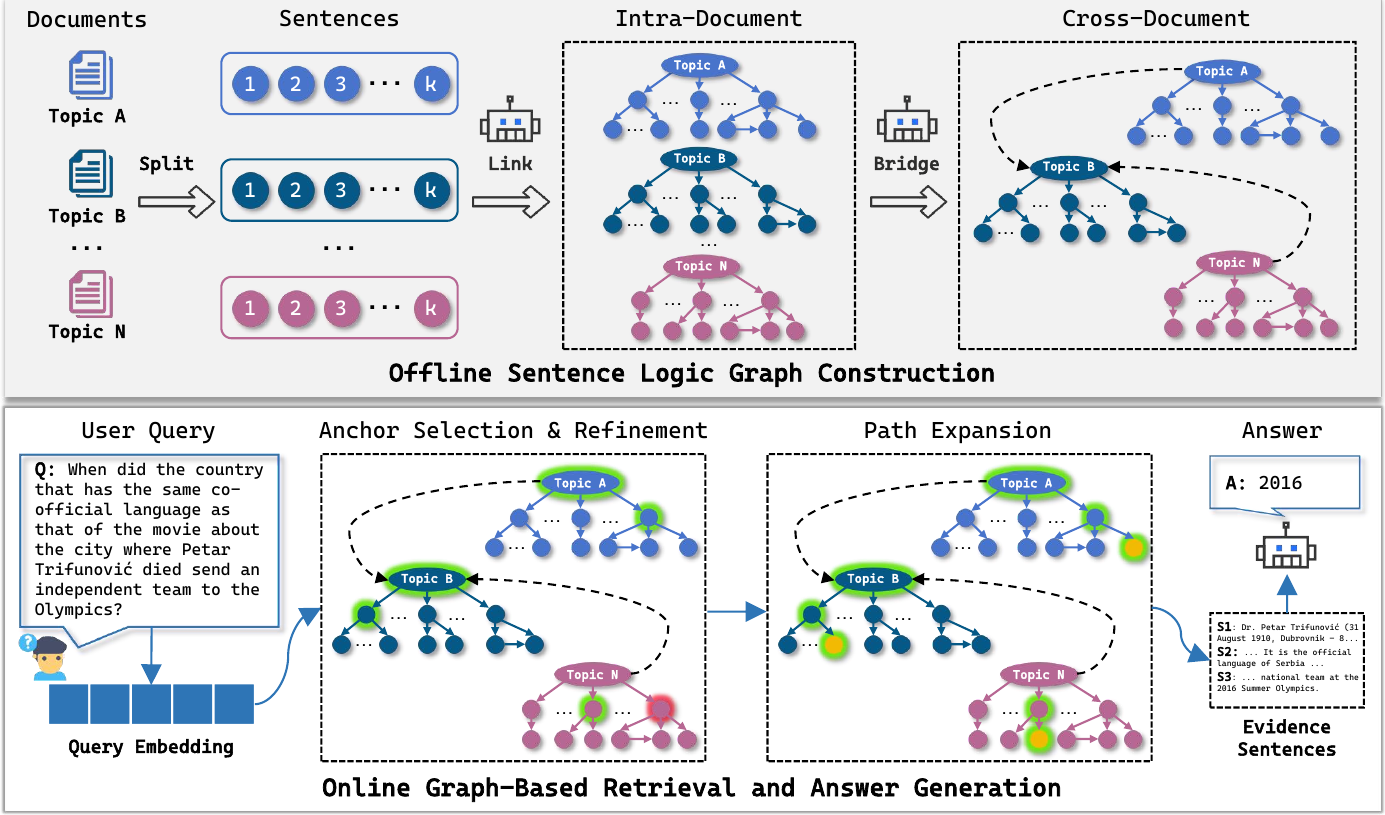}
\caption{An overview of the SentGraph framework. The offline stage constructs a hierarchical sentence logic graph, and the online stage performs graph-based retrieval and answer generation for multi-hop question answering.}
\label{fig2}
\end{figure*}

SentGraph is a sentence-level logic graph construction and retrieval-augmented generation framework for multi-hop question answering. 
Unlike traditional chunk-based retrieval paradigms, SentGraph explicitly captures sentence-level logical dependencies within and across documents. This design enables finer-grained knowledge organization and reasoning path modeling.
As shown in Figure~\ref{fig2}, the framework consists of two stages: offline sentence logic graph construction and online graph-based retrieval and answer generation. In the offline stage, we construct a hierarchical sentence logic graph by modeling sentence-level logical structures. In the online stage, we perform evidence retrieval and answer generation based on the graph, thereby improving LLMs’ effectiveness for cross-document reasoning on complex questions.

\subsection{Offline Stage: Hierarchical Sentence Logic Graph Construction}

Given a document collection, denoted as $D=\{d_1, d_2, \dots, d_n\}$, we first decompose each document into sentence-level semantic units. Compared with traditional chunk-level modeling, sentence-level units provide finer-grained evidence granularity, but also introduce several challenges, including context loss, complex relation modeling, and high computational overhead. To address these challenges, we propose a hierarchical graph construction method based on refined Rhetorical Structure Theory (RST). This method alleviates context loss and relation complexity through RST-based logical relation modeling, while reducing computational overhead through hierarchical architecture design.

Considering that RST defines many relation types, some of which have limited discriminative power for reasoning and QA, we simplify the relation set from the perspective of reasoning function and evidence organization, retaining only relation types that substantially impact cross-document reasoning. Specifically, based on the functional roles of sentences in discourse, we define two categories of relations: \emph{\textbf{Nucleus-Nucleus (N-N) relations}} model logical connections between sentences of equal importance that jointly convey core document semantics, including conjunction, contrast, disjunction, multinuclear restatement, and sequence relations, whereas \emph{\textbf{Nucleus-Satellite (N-S) relations}} model asymmetric dependencies between core sentences and their supporting sentences, including cause, result, opposition, elaboration, circumstance, evaluation, and solutionhood relations.

Based on the above relation modeling, we design a three-layer graph structure that balances expressive power and computational complexity through hierarchical organization. The graph is formally defined as:
$$G=(V,E)$$
where the node set is:
$$V = V_t \cup V_c \cup V_s$$

Here, topic nodes $V_t$ represent document-level topic entities used for cross-document bridging, which are derived from document titles or other document-level metadata. Core sentence nodes $V_c$ correspond to sentences carrying key facts and reasoning support, and supplementary sentence nodes $V_s$ represent subordinate sentences that elaborate on or conditionally supplement core sentences.

The edge set is formally defined as:
$$E = E_{tt} \cup E_{tc} \cup E_{cc} \cup E_{cs}$$

Here, inter-topic bridging edges $E_{tt}$ establish cross-document semantic connections, topic-to-core edges $E_{tc}$ associate topics with their subordinate core sentences, core-to-core edges $E_{cc}$ represent N-N relations such as parallel, contrast, and sequence, and core-to-supplementary edges $E_{cs}$ represent N-S dependencies such as cause, background, and evaluation.

The construction process consists of two steps: intra-document logic modeling and cross-document semantic bridging.

In the intra-document logic modeling stage, we first employ LLMs guided by refined RST to identify core sentences carrying main facts and determine N-N logical relations between them. We then cluster non-core sentences and assign them to corresponding core sentences based on semantic similarity and contextual distance, before using LLMs to establish N-S subordinate relations. This process captures the hierarchical logical structure within each document. Detailed prompt templates are provided in Appendix~\ref{prompt1} and Appendix~\ref{prompt2}.

In the cross-document semantic bridging stage, we identify relations between topic entities across different documents using LLM-guided open information extraction. We then establish inter-topic bridging edges $E_{tt}$ between topic nodes. This enables the reasoning process to integrate key evidence across documents and form cross-document reasoning chains. The detailed prompt template is provided in Appendix~\ref{prompt3}.

\subsection{Online Stage: Graph-Based Retrieval and Answer Generation}

Given a user query, the online process consists of three modules: anchor selection and refinement, adaptive path expansion, and answer generation.

\textbf{Anchor selection and refinement.}
We adopt a coarse-to-fine two-stage strategy. First, a retriever computes similarity scores between the query and all graph nodes, and the Top-$K$ highest-scoring nodes are selected as candidate anchors. These candidates are then refined by the LLM, which filters out loosely related nodes and evaluates whether the remaining anchors contain sufficient evidence. If the evidence is sufficient, the process proceeds directly to answer generation. Otherwise, the process triggers path expansion to retrieve additional evidence. The detailed prompt template used for anchor refinement is provided in Appendix~\ref{prompt4}.

\textbf{Adaptive path expansion.}
We explore reasoning paths starting from each anchor using a breadth-first strategy. For each anchor, we maintain a path queue and expand paths along graph edges by selecting neighboring nodes based on similarity. Newly selected nodes are appended to the current path until a predefined maximum path length or expansion limit is reached.

\textbf{Answer generation.}
We extract all sentence nodes from the retained paths to form the final evidence set. Then, we provide this evidence along with the query to the LLM, instructing it to generate a final answer based on the given context. The LLM performs multi-hop reasoning over the evidence and generates the final answer.

\section{Experiment}

\subsection{Experiment Setting}

We evaluate our method on four multi-hop QA datasets: HotpotQA~\cite{HotpotQA}, 2WikiMultiHopQA (2Wiki)~\cite{2wiki}, MuSiQue~\cite{MuSiQue}, and MultiHopRAG~\cite{multihop}. 
We use Exact Match (EM \%) and F1 score (\%) as the primary evaluation metrics. 
For MultiHopRAG, we report Accuracy following its original evaluation protocol. 
Additionally, we report the retrieval unit type and the average number of retrieved units used by each method.
Further experimental details are provided in Appendix~\ref{app:exp_details}.

\begin{table*}[htpb]
\small
\centering
\setlength{\tabcolsep}{2mm}
\begin{tabular}{cl|lc|ccccccc}
\hline
\multirow{2}{*}{\textbf{Retrieval}} & \multicolumn{1}{c|}{\multirow{2}{*}{\textbf{Model}}} & \multirow{2}{*}{\textbf{\begin{tabular}[c]{@{}c@{}}Retrieval \\ Units\end{tabular}}} & \multirow{2}{*}{\textbf{\begin{tabular}[c]{@{}c@{}}Avg \\ \# Units\end{tabular}}} & \multicolumn{2}{c}{\textbf{HotpotQA}} & \multicolumn{2}{c}{\textbf{2Wiki}} & \multicolumn{2}{c}{\textbf{MuSiQue}} & \textbf{MultiHop} \\
                                    & \multicolumn{1}{c|}{}                                &                                                                                      &                                                                                   & EM                & F1                & EM               & F1              & EM                & F1               & Accuracy          \\ \hline
\multirow{8}{*}{BM25}               & Retrieval Only                                       & Passage                                                                              & 3.00                                                                              & 9.20              & 15.20             & 3.40             & 9.65            & 1.40              & 4.27             & 37.20             \\
                                    & Retrieval Only                                       & Sentence                                                                             & 3.00                                                                              & 31.20             & 42.44             & 23.60            & 29.29           & 9.60              & 16.86            & 61.60             \\
                                    & RankLlama                                            & Passage                                                                              & 5.00                                                                              & 29.48             & 27.82             & 30.30            & 21.91           & 6.04              & 9.26             & 42.09             \\
                                    & SetR-CoT \& IRI                                      & Passage                                                                              & 2.63                                                                              & 32.20             & 30.57             & {\ul 32.17}      & 24.22           & 6.62              & 10.57            & 44.13             \\
                                    
                                    & KGP                                                  & Passage                                                                              & 3.00                                                                              & {\ul 36.82}       & {\ul 49.94}       & 22.80            & 31.21           & 9.31              & 18.66            & {\ul 62.00}       \\
                                    & LightRAG                                             & Passage                                                                              & 3.00                                                                              & 28.66             & 39.06             & 17.78            & 27.43           & {\ul 9.83}        & {\ul 18.71}      & 26.65             \\
                                    & \textbf{SentGraph(Ours)}                             & Sentence                                                                             & 2.57                                                                              & \textbf{43.80}    & \textbf{57.13}    & \textbf{32.20}   & \textbf{39.48}  & \textbf{16.00}    & \textbf{26.83}   & \textbf{63.40}    \\ \hline
\multirow{8}{*}{BGE}                & Retrieval Only                                       & Passage                                                                              & 3.00                                                                              & 11.00             & 18.09             & 3.80             & 10.34           & 2.40              & 6.86             & 44.40             \\
                                    & Retrieval Only                                       & Sentence                                                                             & 3.00                                                                              & 38.60             & 50.88             & 30.80            & 40.01           & 17.00             & 28.57            & 60.20             \\
                                    & RankLlama                                            & Passage                                                                              & 5.00                                                                              & 31.88             & 32.95             & 32.24            & 25.78           & 7.61              & 11.77            & 43.51             \\
                                    & SetR-CoT \& IRI                                      & Passage                                                                              & 2.91                                                                              & 36.62             & 38.11             & 35.44            & 30.35           & 10.79             & 15.43            & 47.14             \\
                                   
                                    & KGP                                                  & Passage                                                                              & 3.00                                                                              & {\ul 44.00}       & {\ul 58.73}       & {\ul 36.80}      & {\ul 48.20}     & {\ul 21.20}       & {\ul 34.72}      & {\ul 63.40}       \\
                                    & LightRAG                                             & Passage                                                                              & 3.00                                                                              & 27.17             & 37.75             & 17.40            & 26.99           & 8.60              & 17.71            & 20.44             \\
                                    & \textbf{SentGraph(Ours)}                             & Sentence                                                                             & 2.70                                                                              & \textbf{48.80}    & \textbf{62.92}    & \textbf{42.00}   & \textbf{52.26}  & \textbf{26.80}    & \textbf{40.36}   & \textbf{65.60}    \\ \hline
\end{tabular}
\caption{Performance (\%) comparison on four multi-hop question answering datasets (passage-level as chunk-level instantiation). Bold and underlined indicate the best and second best performance, respectively.}

\label{tab:main}
\end{table*}

\subsection{Main Results}

\label{results}

We compare SentGraph with baselines on four multi-hop question answering datasets under both sparse (BM25) and dense (BGE) retrieval settings in Table~\ref{tab:main}. SentGraph consistently achieves the best performance across all datasets and retrieval settings. It is important to note that SentGraph operates on sentence-level retrieval units while most baselines use passage-level units, demonstrating that fine-grained evidence modeling with structured logical dependencies is more effective than coarse-grained retrieval for multi-hop QA. Next, we summarize key observations as follows:

\textbf{Fine-grained retrieval is necessary but insufficient.} Sentence-level retrieval significantly outperforms passage-level retrieval in the retrieval-only setting. For instance, under BM25, sentence-level units achieve 31.20 EM on HotpotQA compared to only 9.20 EM for passage-level units, confirming the importance of fine-grained retrieval for multi-hop reasoning. However, sentence-level retrieval alone still substantially underperforms SentGraph, with gaps of 12.6 EM points on HotpotQA and 10.2 points on 2Wiki under BM25. This indicates that modeling logical dependencies between sentences is critical beyond granularity alone.

\begin{table*}[ht]
\small
\centering
\setlength{\tabcolsep}{2.1mm}

\begin{tabular}{cllcccccccc}
\hline
                                     & \multicolumn{1}{c|}{}                                 & \multicolumn{1}{c}{}                                                                                     & \multicolumn{1}{c|}{}                                                                                   & \multicolumn{2}{c}{\textbf{HotpotQA}} & \multicolumn{2}{c}{\textbf{2Wiki}} & \multicolumn{2}{c}{\textbf{MuSiQue}} & \textbf{MultiHop} \\
\multirow{-2}{*}{\textbf{Retrieval}} & \multicolumn{1}{c|}{\multirow{-2}{*}{\textbf{Model}}} & \multicolumn{1}{c}{\multirow{-2}{*}{\textbf{\begin{tabular}[c]{@{}c@{}}Retrieval \\ Unit\end{tabular}}}} & \multicolumn{1}{c|}{\multirow{-2}{*}{\textbf{\begin{tabular}[c]{@{}c@{}}Avg \\ \# Units\end{tabular}}}} & EM                & F1                & EM               & F1              & EM                & F1               & Accuracy          \\ \hline
\multicolumn{11}{c}{\cellcolor[HTML]{EFEFEF}\textbf{Base LLM: Qwen2.5-7B-Instruct}}                                                                                                                                                                                                                                                                                                                                                                       \\
                                     & \multicolumn{1}{l|}{RO}                               & Passage                                                                                                  & \multicolumn{1}{c|}{3.00}                                                                               & 16.60             & 23.31             & 23.80            & 26.43           & 3.20              & 8.16             & 35.80             \\
                                     & \multicolumn{1}{l|}{KGP}                              & Passage                                                                                                  & \multicolumn{1}{c|}{3.00}                                                                               & {\ul 33.20}       & {\ul 46.40}       & {\ul 26.40}      & {\ul 31.55}     & {\ul 6.20}       &  12.15      & {\ul 56.92}       \\
                                     & \multicolumn{1}{l|}{RO}                               & Sentence                                                                                                 & \multicolumn{1}{c|}{3.00}                                                                               & 33.00             & 43.53             & 25.60            & 30.16           & 7.00             & {\ul 14.86}            & 56.60             \\
\multirow{-4}{*}{BM25}               & \multicolumn{1}{l|}{SentGraph}                        & Sentence                                                                                                 & \multicolumn{1}{c|}{2.83}                                                                               & \textbf{45.00}    & \textbf{56.91}    & \textbf{37.20}   & \textbf{44.07}  & \textbf{14.60}    & \textbf{23.94}   & \textbf{63.80}    \\ \hline
                                     & \multicolumn{1}{l|}{RO}                               & Passage                                                                                                  & \multicolumn{1}{c|}{3.00}                                                                               & 15.40             & 22.99             & 24.00            & 26.51           & 3.20              & 8.35             & 36.20             \\
                                     & \multicolumn{1}{l|}{KGP}                              & Passage                                                                                                  & \multicolumn{1}{c|}{3.00}                                                                               & {\ul 42.20}       & {\ul 57.46}       & {\ul 38.80}      & {\ul 47.71}     & 16.00       & {\ul 26.97}      & 58.50       \\
                                     & \multicolumn{1}{l|}{RO}                               & Sentence                                                                                                 & \multicolumn{1}{c|}{3.00}                                                                               & 40.40             & 52.61             & 36.60            & 43.29           & {\ul 16.40}             & 26.78            & {\ul 58.80}             \\
\multirow{-4}{*}{BGE}                & \multicolumn{1}{l|}{SentGraph}                        & Sentence                                                                                                 & \multicolumn{1}{c|}{2.91}                                                                               & \textbf{51.00}    & \textbf{65.47}    & \textbf{46.60}   & \textbf{56.20}  & \textbf{25.60}    & \textbf{36.46}   & \textbf{64.40}    \\ \hline
\multicolumn{11}{c}{\cellcolor[HTML]{EFEFEF}\textbf{Base LLM: Llama-3.1-8B-Instruct}}                                                                                                                                                                                                                                                                                                                                                                     \\
                                     & \multicolumn{1}{l|}{RO}                               & Passage                                                                                                  & \multicolumn{1}{c|}{3.00}                                                                               & 9.20              & 15.20             & 3.40             & 9.65            & 1.40              & 4.27             & 37.20             \\
                                     & \multicolumn{1}{l|}{KGP}                              & Passage                                                                                                  & \multicolumn{1}{c|}{3.00}                                                                               & {\ul 36.82}       & {\ul 49.94}       & 22.80            & {\ul 31.21}     & 9.31              & {\ul 18.66}      & {\ul 62.00}       \\
                                     & \multicolumn{1}{l|}{RO}                               & Sentence                                                                                                 & \multicolumn{1}{c|}{3.00}                                                                               & 31.20             & 42.44             & {\ul 23.60}      & 29.29           & {\ul 9.60}        & 16.86            & 61.60             \\
\multirow{-4}{*}{BM25}               & \multicolumn{1}{l|}{SentGraph}                        & Sentence                                                                                                 & \multicolumn{1}{c|}{2.57}                                                                               & \textbf{43.80}    & \textbf{57.13}    & \textbf{32.20}   & \textbf{39.48}  & \textbf{16.00}    & \textbf{26.83}   & \textbf{63.40}    \\ \hline
                                     & \multicolumn{1}{l|}{RO}                               & Passage                                                                                                  & \multicolumn{1}{c|}{3.00}                                                                               & 11.00             & 18.09             & 3.80             & 10.34           & 2.40              & 6.86             & 44.40             \\
                                     & \multicolumn{1}{l|}{KGP}                              & Passage                                                                                                  & \multicolumn{1}{c|}{3.00}                                                                               & {\ul 44.00}       & {\ul 58.73}       & {\ul 36.80}      & {\ul 48.20}     & {\ul 21.20}       & {\ul 34.72}      & {\ul 63.40}       \\
                                     & \multicolumn{1}{l|}{RO}                               & Sentence                                                                                                 & \multicolumn{1}{c|}{3.00}                                                                               & 38.60             & 50.88             & 30.80            & 40.01           & 17.00             & 28.57            & 60.20             \\
\multirow{-4}{*}{BGE}                & \multicolumn{1}{l|}{SentGraph}                        & Sentence                                                                                                 & \multicolumn{1}{c|}{2.70}                                                                               & \textbf{48.80}    & \textbf{62.92}    & \textbf{42.00}   & \textbf{52.26}  & \textbf{26.80}    & \textbf{40.36}   & \textbf{65.60}    \\ \hline
\multicolumn{11}{c}{\cellcolor[HTML]{EFEFEF}\textbf{Base LLM: Qwen2.5-14B-Instruct}}                                                                                                                                                                                                                                                                                                                                                                      \\
                                     & \multicolumn{1}{l|}{RO}                               & Passage                                                                                                  & \multicolumn{1}{c|}{3.00}                                                                               & 14.00             & 18.91             & 16.20            & 18.26           & 2.20              & 2.42             & 35.60             \\
                                     & \multicolumn{1}{l|}{KGP}                              & Passage                                                                                                  & \multicolumn{1}{c|}{3.00}                                                                               & {\ul 43.66}       & {\ul 56.06}       & {\ul 32.20}      & {\ul 38.37}     & 11.60             & 19.89            & 67.20             \\
                                     & \multicolumn{1}{l|}{RO}                               & Sentence                                                                                                 & \multicolumn{1}{c|}{3.00}                                                                               & 37.20             & 48.94             & 27.00            & 32.08           & {\ul 13.20}       & {\ul 22.10}      & {\ul 66.80}       \\
\multirow{-4}{*}{BM25}               & \multicolumn{1}{l|}{SentGraph}                        & Sentence                                                                                                 & \multicolumn{1}{c|}{2.89}                                                                               & \textbf{49.00}    & \textbf{62.36}    & \textbf{43.80}   & \textbf{51.84}  & \textbf{24.00}    & \textbf{33.93}   & \textbf{69.20}    \\ \hline
                                     & \multicolumn{1}{l|}{RO}                               & Passage                                                                                                  & \multicolumn{1}{c|}{3.00}                                                                               & 13.60             & 18.25             & 10.00            & 10.88           & 3.00              & 4.51             & 39.20             \\
                                     & \multicolumn{1}{l|}{KGP}                              & Passage                                                                                                  & \multicolumn{1}{c|}{3.00}                                                                               & {\ul 53.66}       & {\ul 67.06}       & {\ul 48.20}      & {\ul 56.60}     & {\ul 28.40}       & {\ul 42.37}      & {\ul 69.00}       \\
                                     & \multicolumn{1}{l|}{RO}                               & Sentence                                                                                                 & \multicolumn{1}{c|}{3.00}                                                                               & 43.00             & 56.14             & 40.80            & 49.60           & 22.60             & 34.90            & 66.20             \\
\multirow{-4}{*}{BGE}                & \multicolumn{1}{l|}{SentGraph}                        & Sentence                                                                                                 & \multicolumn{1}{c|}{2.61}                                                                               & \textbf{55.40}    & \textbf{68.74}    & \textbf{54.20}   & \textbf{63.05}  & \textbf{36.80}    & \textbf{49.30}   & \textbf{70.00}    \\ \hline
\multicolumn{11}{c}{\cellcolor[HTML]{EFEFEF}\textbf{Base LLM: Qwen2.5-32B-Instruct}}                                                                                                                                                                                                                                                                                                                                                                      \\
                                     & \multicolumn{1}{l|}{RO}                               & Passage                                                                                                  & \multicolumn{1}{c|}{3.00}                                                                               & 20.80             & 27.78             & 24.00            & 26.88           & 3.40              & 7.75             & 44.80             \\
                                     & \multicolumn{1}{l|}{KGP}                              & Passage                                                                                                  & \multicolumn{1}{c|}{3.00}                                                                               & {\ul 46.08}       & {\ul 58.29}       & {\ul 34.40}      & {\ul 40.20}     & {\ul 14.60}       & {\ul 25.25}      & 66.40             \\
                                     & \multicolumn{1}{l|}{RO}                               & Sentence                                                                                                 & \multicolumn{1}{c|}{3.00}                                                                               & 40.60             & 51.70             & 31.80            & 37.22           & 12.40             & 21.27            & {\ul 67.20}       \\
\multirow{-4}{*}{BM25}               & \multicolumn{1}{l|}{SentGraph}                        & Sentence                                                                                                 & \multicolumn{1}{c|}{2.54}                                                                               & \textbf{51.40}    & \textbf{64.03}    & \textbf{45.80}   & \textbf{53.69}  & \textbf{25.20}    & \textbf{36.78}   & \textbf{72.40}    \\ \hline
                                     & \multicolumn{1}{l|}{RO}                               & Passage                                                                                                  & \multicolumn{1}{c|}{3.00}                                                                               & 21.00             & 28.52             & 23.40            & 26.24           & 5.00              & 9.16             & 51.20             \\
                                     & \multicolumn{1}{l|}{KGP}                              & Passage                                                                                                  & \multicolumn{1}{c|}{3.00}                                                                               & {\ul 55.28}       & {\ul 68.89}       & {\ul 51.80}      & {\ul 60.53}     & {\ul 28.20}       & {\ul 41.41}      & 67.80             \\
                                     & \multicolumn{1}{l|}{RO}                               & Sentence                                                                                                 & \multicolumn{1}{c|}{3.00}                                                                               & 43.20             & 57.01             & 45.00            & 52.76           & 25.00             & 36.89            & {\ul 68.80}       \\
\multirow{-4}{*}{BGE}                & \multicolumn{1}{l|}{SentGraph}                        & Sentence                                                                                                 & \multicolumn{1}{c|}{2.62}                                                                               & \textbf{57.60}    & \textbf{70.64}    & \textbf{55.40}   & \textbf{64.73}  & \textbf{38.80}    & \textbf{52.01}   & \textbf{73.00}    \\ \hline
\end{tabular}

\caption{Performance (\%) comparison across four multi-hop question answering datasets at different LLM scales (passage-level as chunk-level instantiation). ``RO'' denotes retrieval-only. Bold and underlined values indicate the best and second-best results, respectively.}
\label{tab:llms}

\end{table*}

\textbf{Post-retrieval methods show limited gains.} 
RankLlama and SetR-CoT \& IRI improve over passage-level retrieval-only baselines through reranking and result refinement. For instance, under BGE, SetR-CoT \& IRI achieves 36.62 EM on HotpotQA and 35.44 EM on 2WikiMultiHopQA, outperforming the passage-level retrieval-only baseline by 25.62 and 31.64 EM points respectively. However, their effectiveness remains constrained by initial retrieval quality.

\textbf{Graph-enhanced methods benefit from structure but lack fine-grained modeling.} LightRAG and KGP outperform retrieval-only baselines by introducing explicit structural connections. Among all baselines, KGP achieves the strongest performance, reaching 44.00 EM on HotpotQA and 36.80 EM on 2WikiMultiHopQA under BGE retrieval. However, these methods typically construct graphs at the passage level, which limits their ability to capture fine-grained logical relations between sentences. 
In contrast, SentGraph models sentence-level logical dependencies and achieves superior performance. Under BGE setting, SentGraph outperforms KGP by 4.8 EM points on HotpotQA, 5.2 points on 2WikiMultiHopQA, and 5.6 points on MuSiQue. This demonstrates that SentGraph's gains arise from fine-grained evidence selection and structured reasoning paths rather than increased context length.

\subsection{Results across Different Base LLMs}

Table~\ref{tab:llms} presents the performance of SentGraph across multiple base LLMs with different model sizes, ranging from 7B to 32B parameters. We evaluate methods under two key dimensions: retrieval granularity (passage-level vs. sentence-level) and structural modeling (retrieval-only vs. graph-based). SentGraph consistently outperforms all baselines across different model scales, demonstrating the necessity of modeling logical dependencies at the sentence level for multi-hop QA.

Specifically, across all LLM scales, sentence-level retrieval-only consistently outperforms passage-level retrieval-only by substantial margins. For example, with Qwen2.5-32B under BGE retrieval, sentence-level retrieval achieves 43.20 EM on HotpotQA and 45.00 EM on 2WikiMultiHopQA, compared to only 21.00 and 23.40 EM for passage-level retrieval. We also observe that scaling up the LLM generally improves performance across all methods, but the gains for retrieval-only methods remain limited. This suggests that fine-grained retrieval units alone are not enough, and proper evidence organization is still necessary even with stronger LLMs.

The graph-based method KGP improves over passage-level retrieval-only across all LLM scales. With Qwen2.5-32B under BGE retrieval, KGP reaches 55.28 EM on HotpotQA and 51.80 EM on 2WikiMultiHopQA, representing gains of 34.28 and 28.40 EM points over passage-level retrieval-only. This confirms the benefit of explicit structural modeling for multi-hop reasoning. However, coarse-grained passage units inherently mix relevant and irrelevant sentences, limiting the effectiveness of graph-based reasoning. In contrast, SentGraph constructs a hierarchical sentence-level graph that explicitly models logical dependencies between sentences, enabling fine-grained evidence selection and structured reasoning paths. With the same LLM, SentGraph achieves 57.60 EM on HotpotQA, 55.40 EM on 2WikiMultiHopQA, and 38.80 EM on MuSiQue, outperforming sentence-level retrieval-only by 14.4, 10.4, and 13.8 EM points, and surpassing KGP by 2.32, 3.6, and 10.6 EM points respectively. 
Notably, these improvements remain consistent across different LLM scales, demonstrating robust and scalable gains.

\begin{table}[t]
\small
\centering
\setlength{\tabcolsep}{0.9mm}
\begin{tabular}{ccc|cccccc}
\hline
\multicolumn{3}{c|}{\textbf{Components}}   & \multicolumn{2}{c}{\textbf{HotpotQA}} & \multicolumn{2}{c}{\textbf{2Wiki}} & \multicolumn{2}{c}{\textbf{MuSiQue}} \\
AS           & AER          & GPE          & EM                & F1                & EM               & F1              & EM                & F1               \\ \hline
$\checkmark$ & $\times$     & $\times$     & 37.60             & 50.62             & 28.00            & 35.52           & 16.00             & 27.67            \\
$\checkmark$ & $\checkmark$ & $\times$     & 44.80             & 59.37             & 37.40            & 46.96           & 25.40             & 37.57            \\
$\checkmark$ & $\checkmark$ & $\checkmark$ & \textbf{48.80}    & \textbf{62.92}    & \textbf{42.00}   & \textbf{52.26}  & \textbf{26.80}    & \textbf{40.36}   \\ \hline
\end{tabular}
\caption{Ablation study on core components.}
\label{tab:Ablation}
\end{table}

\subsection{Ablation Study}

Table~\ref{tab:Ablation} reports the ablation results of SentGraph by progressively enabling its core components. 
AS denotes anchor selection, AER denotes adaptive evidence refinement, and GPE denotes guided path expansion.
AS alone provides baseline performance by identifying locally relevant sentences, but remains insufficient for complex multi-hop reasoning. 
Introducing AER leads to substantial improvements, with EM gains of 7.2 points on HotpotQA, 9.4 points on 2WikiMultiHopQA, and 9.4 points on MuSiQue, highlighting the importance of filtering irrelevant anchors and assessing evidence sufficiency. Further adding GPE yields additional gains of 4.0, 4.6, and 1.4 EM points, respectively. 
These results demonstrate that SentGraph benefits from the complementary roles of its components, where anchor selection provides initial candidates, evidence refinement filters noise and assesses sufficiency, and path expansion broadens evidence coverage to support multi-hop reasoning.

\subsection{Sentence-level Graph Construction Analysis}

To investigate the effect of sentence-level graph construction and discourse relations, we compare three graph construction strategies:
(1) KGP (Sentence): We adapt the existing graph-based RAG framework KGP by replacing passage-level nodes with sentence-level nodes while keeping its original graph structure.
(2) SentGraph (Rand): We preserve all sentence nodes from SentGraph but replace the RST-based discourse edges with random connections while maintaining the same number of edges.
(3) SentGraph (RST): Our full method with sentence nodes and RST-based discourse relations.

As shown in Table~\ref{tab:rst_ablation}, compared to the passage-level KGP in Table~\ref{tab:main}, adapting KGP to sentence-level nodes results in a significant performance drop across three datasets. This indicates that sentence-level graph construction requires structured discourse modeling rather than a simple change of retrieval granularity.
Furthermore, randomizing the edges also leads to a notable performance drop compared to SentGraph (RST), demonstrating that the improvement stems from explicitly modeled RST discourse relations rather than mere graph connectivity.

\begin{table}[t]
\small
\centering
\setlength{\tabcolsep}{0.7mm}
\begin{tabular}{c|cccccc}
\hline
\multirow{2}{*}{\textbf{Model}} & \multicolumn{2}{c}{\textbf{HotpotQA}} & \multicolumn{2}{c}{\textbf{2Wiki}} & \multicolumn{2}{c}{\textbf{MuSiQue}} \\
                                & EM                & F1                & EM               & F1              & EM                & F1               \\ \hline
KGP (Sentence)                        & 9.74              & 17.42             & 1.81             & 12.15           & 0.81              & 6.04             \\
SentGraph (Rand)               & 41.60             & 54.77             & 34.00            & 42.54           & 17.20             & 29.17            \\
SentGraph (RST)                  & \textbf{48.80}    & \textbf{62.92}    & \textbf{42.00}   & \textbf{52.26}  & \textbf{26.80}    & \textbf{40.36}   \\ \hline
\end{tabular}

\caption{Impact of sentence-level graph construction and RST-based discourse relations on multi-hop QA.}
\label{tab:rst_ablation}
\end{table}

To verify the reliability of automatically generated RST relations, we randomly sample 100 discourse relations and manually annotate them. The Nucleus-Nucleus accuracy is 74\% and Nucleus-Satellite accuracy is 76\%, indicating substantial agreement between LLM predictions and human judgment.

\subsection{Anchor Selection Analysis}

Figure~\ref{fig:anchor_ablation} presents the impact of the number of anchors on SentGraph performance. As the number of anchors increases from 5 to 25, performance consistently improves, with gains of 6.2 EM points on HotpotQA, 8.8 points on 2WikiMultiHopQA, and 8.4 points on MuSiQue. Beyond 20 anchors, the performance improvement becomes smaller. 
This indicates that an adequate pool of candidate anchors is crucial for capturing diverse reasoning paths in multi-hop questions, but excessively large anchor sets yield diminishing returns.

We further evaluate the robustness of the anchor selection module using Recall@K for at least one gold sentence (hit1). On MuSiQue, the module achieves 91.0\% recall at K=5 and 97.2\% recall at K=10. On MultiHopRAG, recall reaches 52.2\% at K=5 and 69.4\% at K=10. These results demonstrate that the anchor selection module consistently identifies gold evidence early, providing reliable entry points for subsequent graph traversal.

\subsection{Performance and Efficiency Analysis}

Figure~\ref{fig:token_usage} compares the token usage of SentGraph and KGP. SentGraph achieves consistent reductions in both input and output token consumption while maintaining superior performance.
For input tokens, SentGraph reduces context length by 29.99\% on HotpotQA, 45.26\% on 2WikiMultiHopQA, and 30.38\% on MuSiQue compared to KGP.
These reductions stem from sentence-level retrieval granularity, which enables more fine-grained evidence selection and helps reduce irrelevant context that is often included in passage-level retrieval.
Output token savings are even more pronounced, with reductions of 69.00\% on HotpotQA, 18.56\% on 2WikiMultiHopQA, and 9.22\% on MuSiQue. This indicates that cleaner input evidence leads to more concise and focused generation.
Combined with the performance improvements shown in Table~\ref{tab:main}, these results demonstrate that SentGraph achieves better accuracy with lower computational cost.

\begin{figure}[t]
\centering
\begin{tikzpicture}
\begin{axis}[
    hide axis,
    xmin=0, xmax=1,
    ymin=0, ymax=1,
    legend style={
        draw=none,
        legend columns=3,
        /tikz/every even column/.append style={column sep=0.14cm},
        font=\small
    },
    legend pos=north west
]
\addlegendimage{color=datasetA, mark=o, mark size=2pt, line width=1pt}
\addlegendentry{HotpotQA}
\addlegendimage{color=datasetB, mark=square, mark size=2pt, line width=1pt}
\addlegendentry{2WikiMultiHopQA}
\addlegendimage{color=datasetC, mark=star, mark size=2.5pt, line width=1pt}
\addlegendentry{MuSiQue}
\end{axis}
\end{tikzpicture}

\begin{subfigure}[b]{0.49\columnwidth}
\centering
\begin{tikzpicture}
\begin{axis}[
    width=1.12\textwidth,
    height=4cm,
    xlabel={Anchor Number},
    ylabel={EM (\%)},
    xlabel style={font=\scriptsize, yshift=0.1cm},
    ylabel style={font=\scriptsize, yshift=-0.15cm},
    xmin=1, xmax=34,
    ymin=15, ymax=55,
    xtick={5,10,15,20,25,30},
    ytick={15,20,25,30,35,40,45,50,55},
    tick label style={font=\scriptsize},
    ymajorgrids=true,
    grid style=dashed,
    every axis plot/.append style={thick}
]

\addplot[
    color=datasetA,
    mark=o,
    mark size=2pt,
    line width=1pt
    ]
    coordinates {
    (5,43.20)(10,47.20)(15,47.40)(20,48.80)(25,49.40)(30,48.60)
    };

\addplot[
    color=datasetB,
    mark=square,
    mark size=2pt,
    line width=1pt
    ]
    coordinates {
    (5,33.40)(10,37.60)(15,40.20)(20,42.00)(25,42.20)(30,43.40)
    };

\addplot[
    color=datasetC,
    mark=star,
    mark size=2.5pt,
    line width=1pt
    ]
    coordinates {
    (5,20.60)(10,25.80)(15,26.20)(20,26.80)(25,29.00)(30,29.20)
    };

\end{axis}
\end{tikzpicture}
\caption*{\hspace*{0.3cm}\parbox{0.9\textwidth}{ (a) EM scores across different anchor numbers.}}
\label{fig:em_scores}
\end{subfigure}
\hfill
\begin{subfigure}[b]{0.49\columnwidth}
\centering
\begin{tikzpicture}
\begin{axis}[
    width=1.12\textwidth,
    height=4cm,
    xlabel={Anchor Number},
    ylabel={F1 (\%)},
    xlabel style={font=\scriptsize, yshift=0.1cm},
    ylabel style={font=\scriptsize, yshift=-0.15cm},
    xmin=1, xmax=34,
    ymin=30, ymax=70,
    xtick={5,10,15,20,25,30},
    ytick={30,35,40,45,50,55,60,65,70},
    tick label style={font=\scriptsize},
    ymajorgrids=true,
    grid style=dashed,
    every axis plot/.append style={thick}
]

\addplot[
    color=datasetA,
    mark=o,
    mark size=2pt,
    line width=1pt
    ]
    coordinates {
    (5,55.90)(10,60.64)(15,61.33)(20,62.92)(25,63.50)(30,63.47)
    };

\addplot[
    color=datasetB,
    mark=square,
    mark size=2pt,
    line width=1pt
    ]
    coordinates {
    (5,43.52)(10,47.51)(15,49.58)(20,52.26)(25,52.92)(30,52.66)
    };

\addplot[
    color=datasetC,
    mark=star,
    mark size=2.5pt,
    line width=1pt
    ]
    coordinates {
    (5,33.30)(10,39.46)(15,39.74)(20,40.36)(25,42.52)(30,42.17)
    };

\end{axis}
\end{tikzpicture}
\caption*{\hspace*{0.15cm}\parbox{0.85\textwidth}{ (b) F1 scores across different anchor numbers.}}

\label{fig:f1_scores}
\end{subfigure}
\caption{Performance across multi-hop question answering datasets with varying anchor numbers.}
\label{fig:anchor_ablation}
\end{figure}

\begin{figure}[htbp]
\centering
\begin{tikzpicture}
\begin{axis}[
    hide axis,
    xmin=0, xmax=1,
    ymin=0, ymax=1,
    legend style={
        draw=none,
        legend columns=2,
        /tikz/every even column/.append style={column sep=2.7cm},
        font=\small,
        legend image code/.code={
            \draw[#1] (0cm,-0.1cm) rectangle (0.3cm,0.1cm);
        }
    },
    legend pos=north west
]
\addlegendimage{fill=datasetA, draw=black, line width=0.5pt}
\addlegendentry{KGP}
\addlegendimage{fill=datasetB, draw=black, line width=0.5pt}
\addlegendentry{SentGraph}
\end{axis}
\end{tikzpicture}

\begin{subfigure}[b]{0.494\columnwidth}
\centering
\begin{tikzpicture}
\begin{axis}[
    width=1.12\textwidth,
    height=4cm,
    ylabel={Input Tokens},
    ylabel style={font=\scriptsize, yshift=-0.15cm},
    ymin=0, ymax=400,
    xtick=data,
    xticklabels={HotpotQA, 2Wiki, MuSiQue},
    x tick label style={font=\scriptsize},
    tick label style={font=\scriptsize},
    xtick pos=left,
    ytick pos=left,
    ymajorgrids=true,
    grid style=dashed,
    ybar=2pt,
    bar width=8pt,
    enlarge x limits=0.25,
    symbolic x coords={HotpotQA, 2Wiki, MuSiQue}
]

\addplot[fill=datasetA, draw=black, line width=0.5pt] coordinates {
    (HotpotQA,292.38) (2Wiki,347.29) (MuSiQue,281.37)
};

\addplot[fill=datasetB, draw=black, line width=0.5pt] coordinates {
    (HotpotQA,204.69) (2Wiki,190.10) (MuSiQue,195.89)
};

\end{axis}
\end{tikzpicture}
\caption*{\hspace*{0.3cm}\parbox{0.85\textwidth}{ (a) Average input token usage per query.}}
\label{fig:input_tokens}
\end{subfigure}
\hfill
\begin{subfigure}[b]{0.494\columnwidth}
\centering
\begin{tikzpicture}
\begin{axis}[
    width=1.12\textwidth,
    height=4cm,
    ylabel={Output Tokens},
    ylabel style={font=\scriptsize, yshift=-0.15cm},
    ymin=0, ymax=35,
    xtick=data,
    xticklabels={HotpotQA, 2Wiki, MuSiQue},
    x tick label style={font=\scriptsize},
    tick label style={font=\scriptsize},
    xtick pos=left,
    ytick pos=left,
    ymajorgrids=true,
    grid style=dashed,
    ybar=2pt,
    bar width=8pt,
    enlarge x limits=0.25,
    symbolic x coords={HotpotQA, 2Wiki, MuSiQue}
]

\addplot[fill=datasetA, draw=black, line width=0.5pt] coordinates {
    (HotpotQA,22.84) (2Wiki,10.29) (MuSiQue,30.66)
};

\addplot[fill=datasetB, draw=black, line width=0.5pt] coordinates {
    (HotpotQA,7.08) (2Wiki,8.38) (MuSiQue,27.83)
};

\end{axis}
\end{tikzpicture}
\caption*{\hspace*{0.15cm}\parbox{0.85\textwidth}{ (b) Average output token usage per query.}}
\label{fig:output_tokens}
\end{subfigure}
\caption{Efficiency analysis on average token usage per query across multi-hop question answering datasets.}
\label{fig:token_usage}
\end{figure}

\section{Conclusion}

We propose SentGraph, a sentence-level graph-based RAG framework for multi-hop QA that constructs hierarchical sentence graphs with explicit logical dependencies by adapting RST. SentGraph further employs a graph-guided retrieval strategy to enable fine-grained evidence selection at the sentence level. Extensive experiments show that SentGraph achieves consistent performance improvements with lower token consumption, highlighting the importance of fine-grained logical dependency modeling for effective multi-hop QA.

\section*{Limitations}

While SentGraph demonstrates strong performance on multi-hop question answering, it currently operates purely on textual data and cannot leverage multimodal information such as images and tables.  In many real-world domains, including scientific literature, technical manuals, and medical documents, there exists critical multimodal evidence that provides substantial practical value. Modeling and reasoning over such multimodal evidence is essential for broader applicability.

In future work, we could explore integrating information from text, images, and structured data into a unified graph representation. We could also design reasoning strategies that can effectively traverse and combine evidence across modalities. These efforts aim to extend SentGraph to incorporate multimodal sources and develop an effective multimodal RAG framework.

\section*{Ethics Statement}

This work focuses on improving multi-hop retrieval and reasoning through structured sentence-level representations. The proposed method is evaluated on publicly available benchmark datasets and does not involve the collection or use of personal, sensitive, or proprietary data. SentGraph does not train or modify LLMs. It operates as a retrieval framework that provides structured context to existing LLMs for answer generation. Like other RAG frameworks, ethical considerations depend on downstream application contexts and require appropriate safeguards during deployment.

\bibliography{rag}

\appendix

\section{Experiment Setting Details}

\label{app:exp_details}
\stitle{Datasets.}
We evaluate our approach on four complex multi-hop question answering datasets.
HotpotQA~\cite{HotpotQA} contains 113k question-answer pairs that require cross-document reasoning and provide sentence-level supporting facts.
2WikiMultiHopQA (2Wiki)~\cite{2wiki} is a multi-hop question answering dataset constructed from Wikipedia, requiring reasoning across multiple documents.
MuSiQue~\cite{MuSiQue} contains approximately 25k questions and enforces multi-step reasoning with 2--4 hops to avoid shortcut solutions.
MultiHopRAG(MultiHop)~\cite{multihop} is a benchmark designed for RAG systems and includes 2,556 questions whose answers must be synthesized from 2--4 news articles.

\stitle{Source Models.} 
We evaluate our method using two retrieval models and multiple LLMs.
For retrieval, we use BM25~\cite{bm25}, a traditional unsupervised method, and bge-large-en-v1.5(BGE)~\cite{bge}, a supervised dense retriever.
For generation, we use Llama-3.1-8B-Instruct~\cite{llama3.1} as the primary LLM.
To evaluate generalization across model scales, we additionally test Qwen2.5-7B-Instruct, Qwen2.5-14B-Instruct, and Qwen2.5-32B-Instruct~\cite{qwen2,qwen2.5}.

Unless otherwise specified, all experiments including ablation studies and analysis use Llama-3.1-8B-Instruct~\cite{llama3.1} with BGE~\cite{bge} retrieval as the default configuration.

\stitle{Baselines.}
(1) \textbf{Retrieval Only(RO)} serves as a basic baseline and follows the standard in-context RAG paradigm~\cite{ragcontext}.
We evaluate both passage-level and sentence-level retrieval units, where the sentence-level baseline uses simple document sentence splitting.
(2) \textbf{RankLlama}~\cite{RankLlama} and \textbf{SetR-CoT \& IRI}~\cite{Shifting} are post-retrieval optimization methods.
(3) \textbf{KGP}~\cite{KGP} and \textbf{LightRAG}~\cite{lightrag} are graph-enhanced RAG methods that incorporate graph structures for multi-hop QA.

Unless otherwise specified, baseline methods follow their original configurations and retrieve passage-level units.
In contrast, our method performs retrieval at the sentence level.

\stitle{Implementation Details.}
All experiments are conducted on a Linux server equipped with four NVIDIA A800 80GB GPUs, dual Intel Xeon CPUs (2.9,GHz), and 512,GB of memory.
Following common practice in prior RAG evaluations, we randomly sample 500 questions from each dataset and use the same evaluation subsets across all methods to ensure fair comparison.
All models are evaluated with deterministic decoding with the temperature set to 0. We repeat each experiment 5 times and report the averaged results to ensure stability.

\section{Prompt}

\label{sec:prompt}

\subsection{Prompt for N-N Relations Recognition}
\label{prompt1}

\begin{figure}[!htbp]
\centering
\includegraphics[width=0.95\columnwidth]{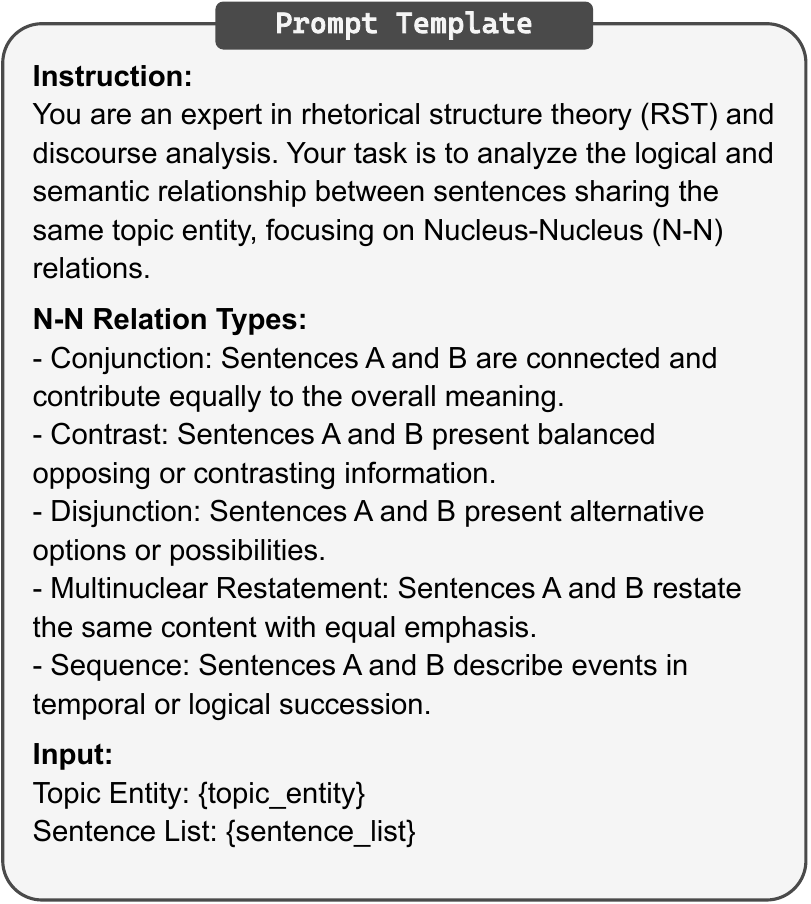}
\caption{Prompt Template for N-N Relations Recognition.}
\label{fig:prompt_nn}
\end{figure}

In the intra-document logic modeling stage, we first identify core sentences and their N–N (Nucleus–Nucleus) logical relations. To accomplish this, we design a prompt that instructs the LLM to recognize five types of N–N relations between sentences. The detailed prompt template is shown in Figure~\ref{fig:prompt_nn}.

\subsection{Prompt for N-S Relations Recognition}
\label{prompt2}

\begin{figure}[!htbp]
\centering
\includegraphics[width=0.95\columnwidth]{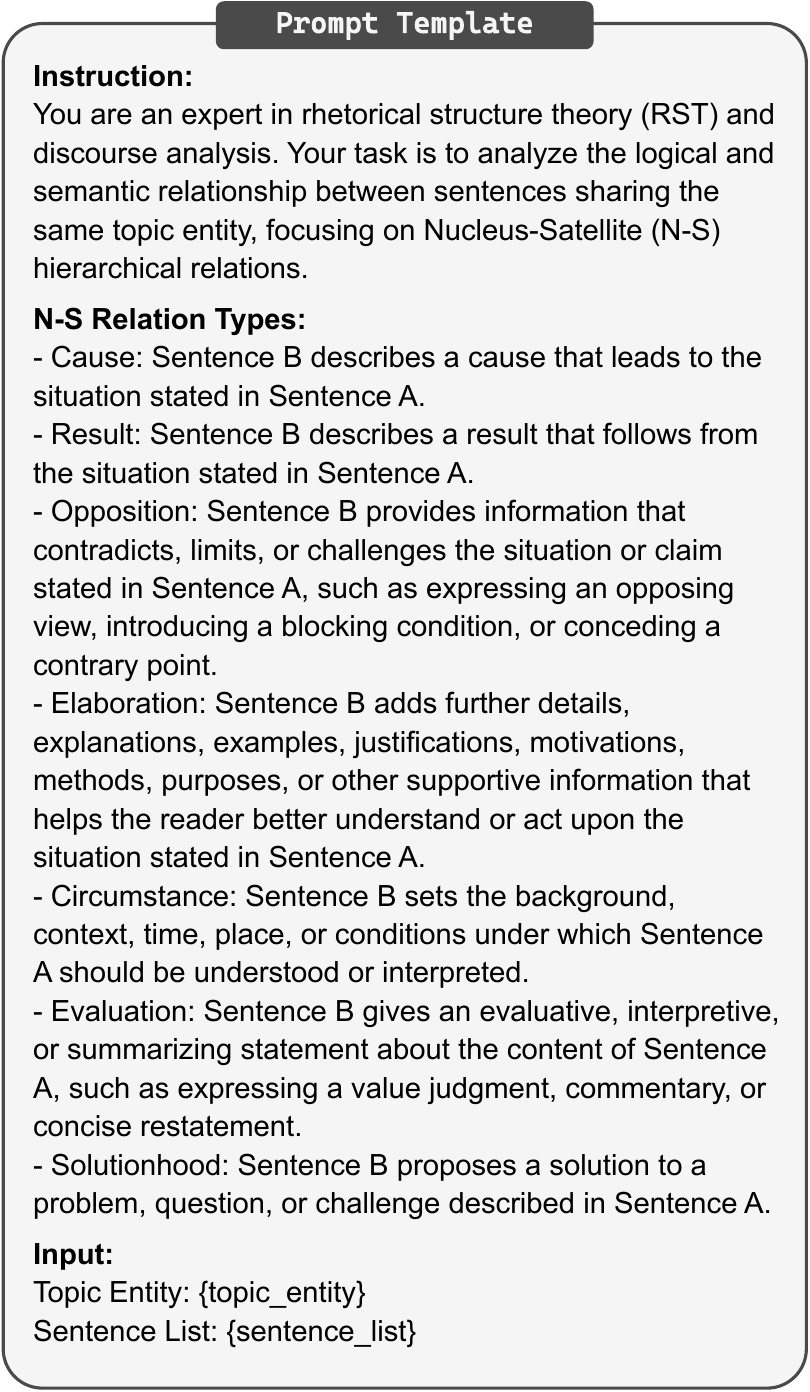}
\caption{Prompt Template for N-S Relations Recognition.}
\label{fig:prompt_ns}
\end{figure}

After identifying core sentences, we cluster non-core sentences and establish N-S (Nucleus-Satellite) relations between them. We design a prompt that instructs the LLM to recognize seven types of N-S relations. The detailed prompt template is shown in Figure~\ref{fig:prompt_ns}.

\subsection{Prompt for Cross-document Semantic Bridging}
\label{prompt3}

\begin{figure}[!htbp]
\centering
\includegraphics[width=0.95\columnwidth]{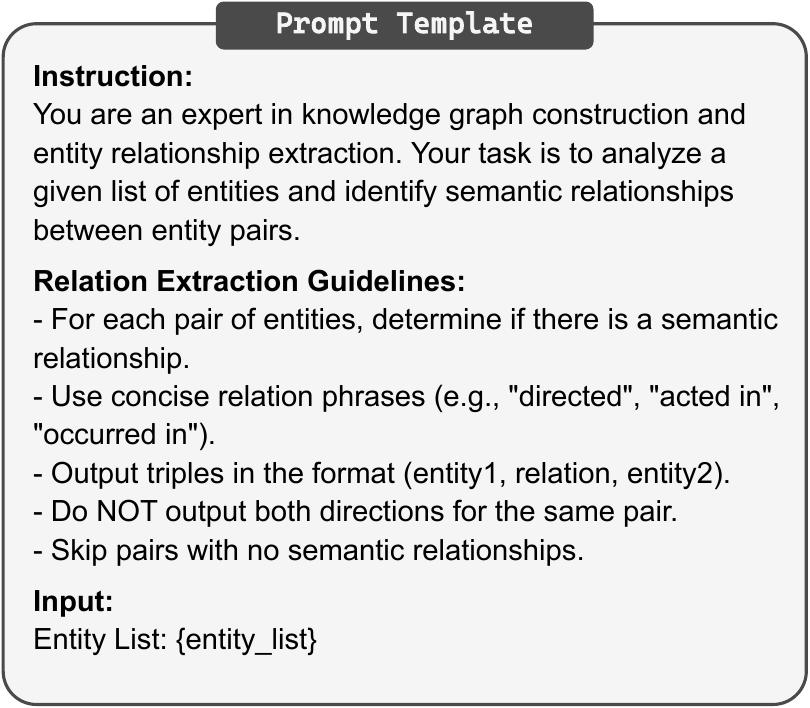}
\caption{Prompt Template for Cross-document Semantic Bridging.}
\label{fig:prompt_bridging}
\end{figure}

To connect information across different documents, we identify semantic relations between topic entities that are grounded in the input corpus. We design a prompt that instructs the LLM to perform corpus-grounded open information extraction over topic entities. The detailed prompt template is shown in Figure~\ref{fig:prompt_bridging}.

\subsection{Prompt for Anchor Refinement}
\label{prompt4}
\begin{figure}[!htbp]
\centering
\includegraphics[width=0.95\columnwidth]{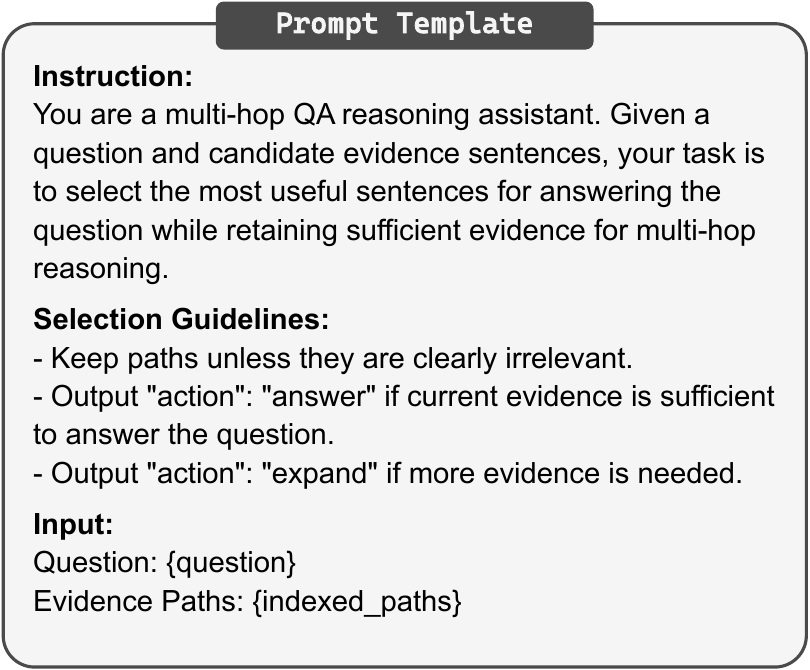}
\caption{Prompt Template for Anchor Refinement.}
\label{fig:prompt_anchor}
\end{figure}

In the online reasoning stage, we refine the initially retrieved anchor nodes. We design a prompt that instructs the LLM to evaluate which evidence paths are most relevant for answering the given query. The detailed prompt template is shown in Figure~\ref{fig:prompt_anchor}.

\end{document}